\documentclass[10pt,twocolumn,letterpaper]{article}

\usepackage{cvpr}              
\usepackage[accsupp]{axessibility}  

\usepackage{amsfonts}       
\usepackage{nicefrac}       

\usepackage{graphicx}
\usepackage{comment}
\usepackage{amsmath,amssymb,amstext} 

\usepackage{gensymb}

\usepackage{relsize}
\usepackage[T1]{fontenc}
\usepackage[scaled=0.90]{inconsolata}
\usepackage[latin1]{inputenc}
\usepackage[english]{babel}

\usepackage{epsfig}
\usepackage{graphicx}
\usepackage{wrapfig}
\usepackage[belowskip=2pt,labelfont=bf,aboveskip=0pt,font=small,labelsep=period]{caption}
\usepackage{array}

\usepackage[belowskip=0pt,aboveskip=0pt,font=small]{subcaption}
\usepackage{longtable}
\usepackage{microtype}

\usepackage{textcomp}
\usepackage{stmaryrd}
\SetSymbolFont{stmry}{bold}{U}{stmry}{m}{n}
\usepackage{upgreek}
\usepackage{bm}
\usepackage{cases}
\usepackage{mathtools}
\usepackage{arydshln}
\usepackage{multirow}

\usepackage{color}
\usepackage[dvipsnames]{xcolor}
\definecolor{JalapenoRed}{RGB}{183,21,64}
\definecolor{Belize}{RGB}{41,128,185}
\definecolor{Amour}{RGB}{238,82,83}
\usepackage{colortbl}

\usepackage[pagebackref=true,breaklinks=true,colorlinks,bookmarks=false,allcolors=JalapenoRed]{hyperref}

\usepackage[capitalize]{cleveref}
\crefname{section}{Sec.}{Secs.}
\Crefname{section}{Section}{Sections}
\Crefname{table}{Table}{Tables}
\crefname{table}{Tab.}{Tabs.}

\usepackage{algorithm}
\usepackage{algpseudocode}

\usepackage{multirow}
\usepackage{rotating}
\usepackage{booktabs}
\usepackage{etoolbox}
\newcounter{magicrownumbers}
\preto\tabular{\setcounter{magicrownumbers}{0}}
\newcommand\rownumber{\stepcounter{magicrownumbers}\arabic{magicrownumbers})\,}

\usepackage{enumitem}
\usepackage[olditem,oldenum]{paralist}

\usepackage{alltt}
\usepackage{listings}

\usepackage{url}
\usepackage{xspace}
\usepackage{comment}
\usepackage{cuted}
\usepackage{caption}

\usepackage{quoting}
\quotingsetup{vskip=0pt}
\usepackage{epigraph}
\usepackage[normalem]{ulem}

\setdefaultleftmargin{.2cm}{.2cm}{}{}{}{}
\setlist{leftmargin=.2cm}

\usepackage{mysymbols}

\graphicspath{{images/}}

\usepackage{graphicx}
\usepackage{amsmath}
\usepackage{amssymb}
\usepackage{booktabs}
\usepackage{tablefootnote}

\newcolumntype{=}{
  >{\gdef\@rowstyle{}}%
}

\newcolumntype{+}{
  >{\@rowstyle}%
}

\usepackage[capitalize]{cleveref}
\crefname{section}{Sec.}{Secs.}
\Crefname{section}{Section}{Sections}
\Crefname{table}{Table}{Tables}
\crefname{table}{Tab.}{Tabs.}

\def\cvprPaperID{3969} 
\def\confName{CVPR}
\def\confYear{2023}

\begin{document}

\title{PIRLNav: Pretraining with Imitation and RL Finetuning for \objnav}

\author{
  Ram Ramrakhya$^1$ \,\,
  Dhruv Batra$^{1,2}$ \,\,
  Erik Wijmans$^1$ \,\,
  Abhishek Das$^{2}$ \vspace{3pt}\\
  $^1$Georgia Institute of Technology \quad $^2$FAIR, Meta AI\\
  {\tt\small $^1$\{ram.ramrakhya,dbatra,etw\}@gatech.edu}
  \quad {\tt\small $^2$abhshkdz@meta.com}
}

\maketitle

\begin{abstract}
We study \objnavfull~-- where a virtual robot
situated in a new environment is asked to navigate to an object.
Prior work~\cite{ramrakhya2022} has shown that imitation
learning (IL) using behavior cloning (BC)
on a dataset of human demonstrations achieves promising
results.
However, this has limitations -- 1) BC policies
generalize poorly to new states, since the training
mimics actions not their consequences, and 2) collecting
demonstrations is expensive.
On the other hand, reinforcement learning (RL) is trivially scalable,
but requires careful reward engineering to achieve desirable behavior.
We present PIRLNav, a two-stage learning scheme for BC pretraining
on human demonstrations followed by RL-finetuning.
This leads to a policy that achieves a success rate of $65.0\%$
on \objnav ($+5.0\%$ absolute over previous state-of-the-art).

%

Using this BC$\rightarrow$RL training recipe, we present a rigorous
empirical analysis of design choices. First, we investigate whether
human demonstrations can be replaced with `free' (automatically generated) sources of demonstrations,~\eg
shortest paths (SP) or task-agnostic frontier exploration (FE) trajectories.
%
We find that BC$\rightarrow$RL on human demonstrations outperforms
BC$\rightarrow$RL on SP and FE trajectories, even when
controlled for the same BC-pretraining success on \textsc{train}, and
even on a subset of \textsc{val} episodes where BC-pretraining success favors
the SP or FE policies.
Next, we study how RL-finetuning performance scales with the size of the BC pretraining dataset.
We find that as we increase the size of the BC-pretraining dataset and get to
high BC accuracies, the improvements from RL-finetuning are smaller,
and that $90\%$ of the performance of our best BC$\rightarrow$RL policy can be
achieved with less than half the number of BC demonstrations.
Finally, we analyze failure modes of our \objnav policies, and present guidelines for further improving them.\\
Project page: \href{https://ram81.github.io/projects/pirlnav}{\tt{ram81.github.io/projects/pirlnav}}.
\end{abstract}

\section{Introduction}
\label{sec:intro}

\begin{figure}[t]
\centering
    \includegraphics[width=0.99\linewidth]{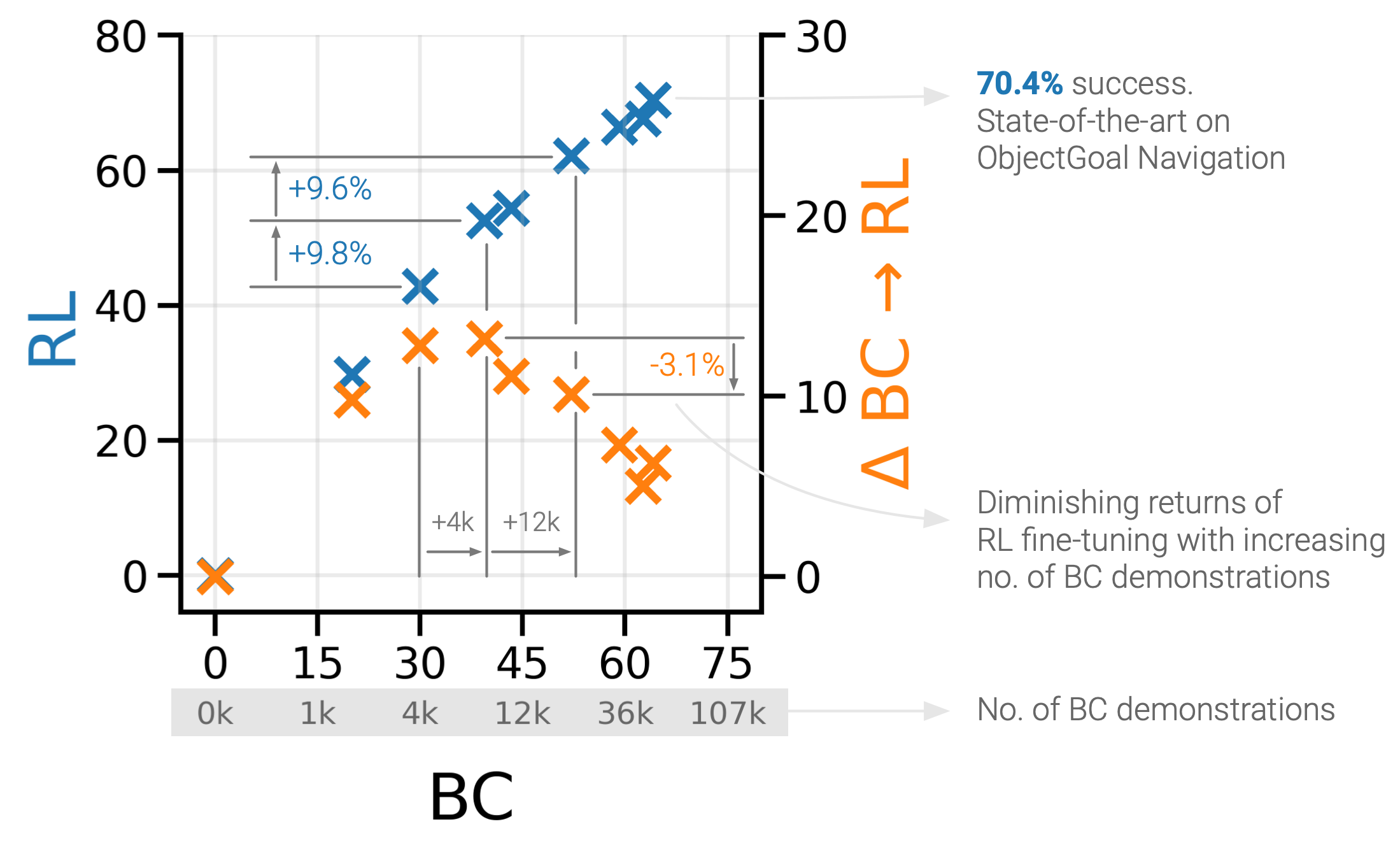}
    \caption{\objnav success rates of agents trained using behavior cloning
      (BC)~\vs BC-pretraining followed by reinforcement learning (RL)
      ({\color{NavyBlue} in blue}). RL from scratch (\ie BC=$0$) fails to get
      off-the-ground. With more BC demonstrations, BC success increases, and it
      transfers to even higher RL-finetuning success. But the difference between
      RL-finetuning~\vs BC-pretraining success ({\color{Orange} in orange})
      plateaus and starts to decrease beyond a certain point, indicating
      diminishing returns with each additional BC demonstration.}
    \label{fig:teaser}
    \vspace{-15pt}
\end{figure}

\begin{figure*}[t]
\centering
    \includegraphics[width=0.98\linewidth]{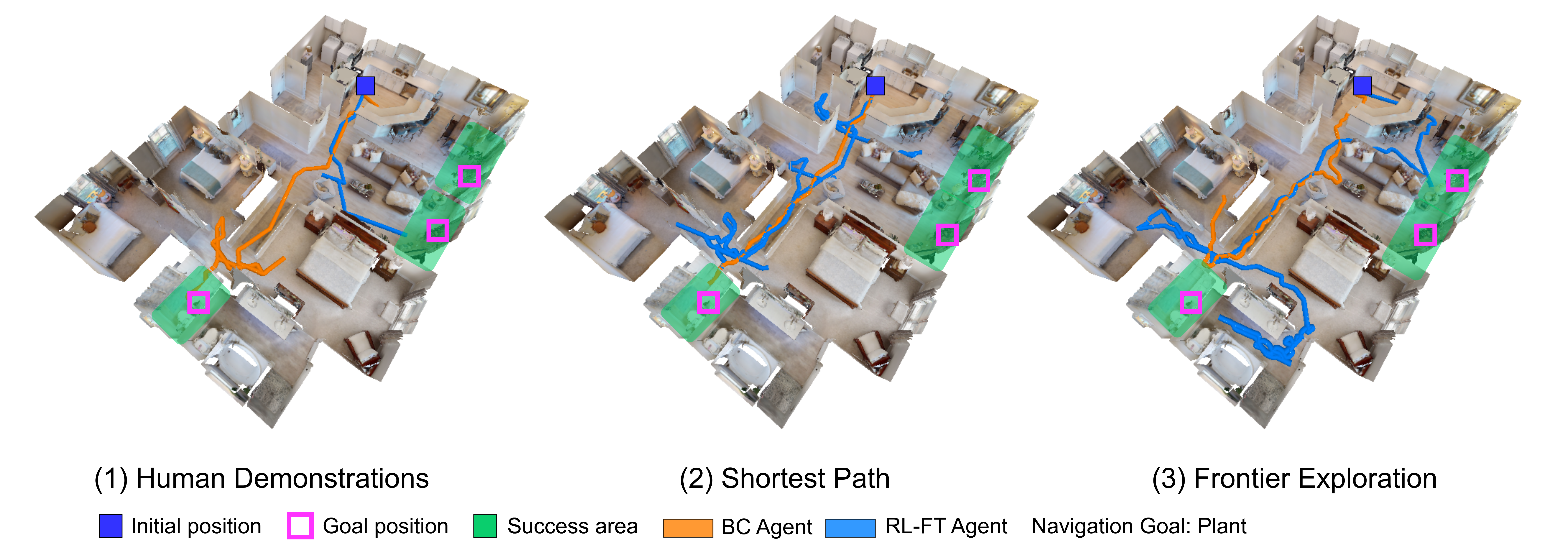}
    \vspace{5pt}
    \caption{\objnav trajectories for policies trained with BC$\rightarrow$RL
        on 1) Human Demonstrations, 2) Shortest Paths, and 3) Frontier Exploration Demonstrations.}
    \label{fig:objnav_il_rl_trajs}
    \vspace{-10pt}
\end{figure*}

Since the seminal work of Winograd~\cite{winograd_cogpsy72}, designing embodied
agents that have a rich understanding of the environment they are situated in,
can interact with humans (and other agents) via language, and the environment via
actions has been a long-term goal in AI~\cite{smith_al05,hermann_arxiv17,hill_arxiv17,chaplot_aaai18,anderson_cvpr18,jain2019two,das_phd_thesis_2020,abramson2020imitating,weihs2021learning,lynch2022interactive}.
We focus on \objnavfull~\cite{anderson_arxiv18,objectnav_tech_report}, wherein an agent situated in a new environment is
asked to navigate to any instance of an object category
(`find a plant', `find a bed',~\etc); see~\figref{fig:objnav_il_rl_trajs}.
\objnav is simple to explain but difficult for today's techniques to accomplish.
First, the agent needs to be able to ground the tokens in the language
instruction to physical objects in the environment
(\eg what does a `plant' look like?).
Second, the agent needs to have rich semantic priors to guide its navigation
to avoid wasteful exploration (\eg the microwave is likely to be found in the
kitchen, not the washroom).
Finally, it has to keep track of where it has been in its internal memory to avoid redundant search.

Humans are adept at \objnav. Prior work~\cite{ramrakhya2022}
collected a large-scale dataset of $80k$ human demonstrations for \objnav,
where human subjects on Mechanical Turk teleoperated virtual robots and searched for objects in novel houses.
This first provided a human baseline on \objnav of $88.9\%$ success rate on the Matterport3D (MP3D) dataset~\cite{mp3d}\footnote{On \textsc{val} split, for 21 object categories, and a maximum of 500 steps.} compared to $35.4\%$ success rate of the best performing method~\cite{ramrakhya2022}.
%
This dataset was then used to train agents via imitation learning (specifically, behavior cloning).

While this approach achieved state-of-art results ($35.4\%$ success rate on MP3D \textsc{val} dataset), it has two clear limitations.
First, behavior cloning (BC) is known to suffer from poor
generalization to out-of-distribution states not seen during training, since the
training emphasizes imitating actions not accomplishing their goals.
Second and more importantly, it is expensive and thus not scalable.
Specifically, Ramrakhya~\etal~\cite{ramrakhya2022} collected
$80k$ demonstrations on $56$ scenes in Matterport3D Dataset, which took ${\sim}2894$ hours of human teleoperation
and \$$50k$ dollars.
A few months after \cite{ramrakhya2022} was released, a new higher-quality dataset called HM3D-Semantics v0.1~\cite{yadav2022habitat} became available
with $120$ annotated 3D scenes, and a few months after that
HM3D-Semantics v0.2 added $96$ additional scenes.
Scaling Ramrakhya \etal's approach to continuously incorporate
new scenes involves replicating
that entire effort again and again.

On the other hand, training with reinforcement learning (RL) is trivially scalable
once annotated 3D scans are available.
However, as demonstrated in Maksymets~\etal\cite{thda_iccv21},
RL requires careful reward engineering, the reward function typically used for \objnav
actually \emph{penalizes} exploration (even though the task requires it),
and the existing RL policies overfit to the small number of
available environments.

Our primary technical contribution is PIRLNav, an approach for pretraining with BC and finetuning with RL for \objnav.
%
BC pretrained policies provide a reasonable starting point for
`bootstrapping' RL and make the optimization easier than learning from scratch.
In fact, we show that BC pretraining even unlocks RL with sparse rewards.
Sparse rewards
are simple (do not involve any reward engineering) and do not suffer from the unintended
consequences described above. However, learning from scratch with
sparse rewards is typically out of reach since most random action trajectories result in no positive rewards.

%

While combining IL and RL has been studied in prior work~\cite{schaal_neurips96,eqa_modular,rajeswaran_RSS_18,vpt22,lynch2019play},
the main technical challenge in the context of modern neural networks is that
imitation pretraining results in weights for the policy (or actor),
but not a value function (or critic).
Thus, naively initializing a new RL policy with these BC-pretrained policy weights
often leads to catastrophic failures due to destructive policy updates
early on during RL training, especially for actor-critic RL methods \cite{jsrl}.
To overcome this challenge, we present a two-stage learning scheme involving
a critic-only learning phase first that gradually
transitions over to training both the actor and critic.
We also identify a set of practical recommendations for this
recipe to be applied to \objnav.
This leads to a PIRLNav policy that advances the
state-the-art on \objnav from $60.0\%$ success rate (in \cite{chaplot_neurips20})
to $65.0\%$ ($+5.0\%$, $8.3\%$ relative improvement).
%

Next, using this BC$\rightarrow$RL training recipe, we
conduct an empirical analysis of design choices.
%
Specifically, an ingredient we investigate is whether
human demonstrations can be replaced with `free' (automatically generated)
sources of demonstrations for \objnav,~\eg (1) shortest paths (SP) between the
agent's start location and the closest object instance, or (2) task-agnostic
frontier exploration~\cite{frontier} (FE) of the environment
followed by shortest path to goal-object upon observing it.
We ask and answer the following:
\begin{compactenum}
\item \myquote{Do human demonstrations capture any unique \objnav-specific behaviors that
  shortest paths and frontier exploration trajectories do not?}
  Yes. We find that BC / BC$\rightarrow$RL on human demonstrations outperforms
  BC / BC$\rightarrow$RL on shortest paths and frontier exploration trajectories
  respectively.
  When we control the number of demonstrations from each source
  such that BC success on \textsc{train} is the same,
  RL-finetuning when initialized from BC on human demonstrations still outperforms
  the other two.
\item \myquote{How does performance after RL scale with BC dataset size?}
We observe diminishing returns from RL-finetuning as we scale BC dataset size.
 This suggests, by effectively leveraging the trade-off curve between size of pretraining
 dataset size \vs performance after RL-Finetuning, we can achieve closer to
 state-of-the-art results without investing into a large dataset of BC demonstrations.
\item \myquote{Does BC on frontier exploration demonstrations present similar
  scaling behavior as BC on human demonstrations?}
No. We find that as we scale frontier exploration demonstrations past
$70k$ trajectories, the performance plateaus.
\end{compactenum}

Finally, we present an analysis of the failure modes of our \objnav policies
and present a set of guidelines for further improving them.
Our policy's primary failure modes are: a) Dataset issues: comprising of missing goal
annotations, and navigation meshes blocking the path,
b) Navigation errors: primarily failure to navigate between floors,
c) Recognition failures: where the agent does not identify the goal object during an episode,
or confuses the specified goal with a semantically-similar object.
\section{Related Work}
\label{sec:related_work}

\textbf{ObjectGoal Navigation}.
Prior works on \objnav have used end-to-end RL~\cite{mousavian2018sem_midlevel,ye_iccv21,thda_iccv21},
modular learning \cite{chaplot_neurips20,liang2020sscnav,ramakrishnan2022poni}, and imitation learning \cite{ramrakhya2022,ovrl}.
Works that use end-to-end RL have proposed improved visual representations \cite{mousavian2018sem_midlevel,wei2019},
auxiliary tasks \cite{ye_iccv21}, and data augmentation techniques \cite{thda_iccv21} to improve generalization to
unseen environments.
Improved visual representations include object relation graphs \cite{wei2019} and
semantic segmentations \cite{mousavian2018sem_midlevel}.
Ye~\etal\cite{ye_iccv21} use auxiliary tasks like predicting environment dynamics, action distributions, and map coverage
in addition to \objnav and achieve promising results.
Maksymets~\etal~\cite{thda_iccv21} improve generalization of RL agents by training
with artificially inserted objects and proposing a reward to incentivize exploration.

Modular learning methods for \objnav have also emerged as a strong competitor \cite{chaplot_neurips20,liang2020sscnav,ramakrishnan_arxiv20}.
These methods rely on separate modules for semantic mapping that build explicit structured map representations,
a high-level semantic exploration module that is learned through RL to solve the `where to look?' subproblem,
and a low-level navigation policy that solves `how to navigate to $(x, y)$?'.

The current state-of-the-art methods on \objnav~\cite{ramrakhya2022,ovrl}
make use of BC on a large dataset of $80k$ human demonstrations.
%
with a simple CNN+RNN policy architecture.
In this work, we improve on them by developing an effective approach
to finetune these imitation-pretrained policies with RL.
%

\textbf{Imitation Learning and RL Finetuning}.
Prior works have considered a special case of learning from demonstration data.
These approaches initialize policies trained using behavior cloning, and then fine-tune
using on-policy reinforcement learning \cite{schaal_neurips96,rajeswaran_RSS_18,vpt22,lynch2019play,jan2008,jens2008},
On classical tasks like cart-pole swing-up \cite{schaal_neurips96}, balance,
hitting a baseball \cite{jan2008}, and underactuated swing-up \cite{jens2008},
demonstrations have been used to speed up learning by initializing policies pretrained on demonstrations
for RL.
Similar to these methods, we also use a on-policy RL algorithm for finetuning the policy trained
with behavior cloning.
Rajeswaran~\etal\cite{rajeswaran_RSS_18} (DAPG) pretrain a policy using behavior cloning
and use an augmented RL finetuning objective to stay close to the demonstrations
which helps reduce sample complexity.
Unfortunately DAPG is not feasible in our setting as it requires solving a systems research problem
to efficiently incorporate replaying demonstrations and collecting experience online at our scale.
\cite{rajeswaran_RSS_18} show results of the approach on a dexterous hand manipulation task with a 
small number of demonstrations that can be loaded in system memory and therefore did not need to solve this
system challenge.
This is not possible in our setting, just the 256$\times$256 RGB observations for the
$77k$ demos we collect would occupy over 2 TB memory, which is out of reach for all but the most
exotic of today's systems.
There are many methods for incorporating demonstrations/imitation learning with off-policy RL~\cite{awac,awopt2021corl,kalashnikov2018scalable,AWRPeng19,marwil}.
Unfortunately these methods were not designed to work with recurrent policies
and adapting off-policy methods to work with recurrent policies is challenging~\cite{r2d2_iclr19}.
See the ~\cref{sec:prior_works} for more details.
%
%
%
The RL finetuning approach that demonstrates results with an actor-critic
and high-dimensional visual observations, and is thus most closely related to our
setup is proposed in VPT~\cite{vpt22}.
Their approach uses Phasic Policy Gradients (PPG)~\cite{cobbe2020ppg}
with a KL-divergence loss between the current policy and the frozen
pretrained policy, and decays the KL loss weight $\rho$ over time to enable exploration
during RL finetuning.
Our approach uses Proximal Policy Gradients (PPO)~\cite{schulman_arxiv17}
instead of PPG, and therefore does not require a KL constraint, which is compute-expensive,
and performs better  on \objnav.
%
%

\section{\objnav and Imitation Learning}
\label{sec:training_details}


\subsection{\objnav}

In \objnav an agent is tasked with searching for an instance of the
specified object category (\eg, `bed') in an unseen environment.
The agent must perform this task using only egocentric perceptions.
Specifically, a RGB camera, Depth sensor\footnote{We don't use this sensor as we don't find it helpful.}, and a GPS+Compass sensor
that provides location and orientation relative to the start position of the
episode.
The action space is discrete and consists of \moveforward ($0.25m$),
\turnleft ($30^{\circ}$), \turnright ($30^{\circ}$), \lookup ($30^{\circ}$),
\lookdown ($30^{\circ}$), and \stopac actions.
An episode is considered successful if the agent stops within $1m$ Euclidean distance of the goal
object within $500$ steps and is able to view the object by taking turn actions~\cite{objectnav_tech_report}.

We use scenes from the HM3D-Semantics v0.1 dataset~\cite{yadav2022habitat}.
The dataset consists of $120$ scenes and $6$ unique goal object categories.
We evaluate our agent using the train/val/test splits from the 2022 Habitat Challenge\footnote{\url{https://aihabitat.org/challenge/2022/}}.

\subsection{\objnav Demonstrations}

Ramrakhya~\etal~\cite{ramrakhya2022} collected \objnav demonstrations for the Matterport3D dataset~\cite{mp3d}.
We begin our study by replicating this effort and collect demonstrations for the HM3D-Semantics v0.1 dataset~\cite{yadav2022habitat}. We use Ramrakhya~\etal's Habitat-WebGL infrastructure to collect $77k$ demonstrations, amounting to ${\sim}2378$ human annotation hours.

\subsection{Imitation Learning from Demonstrations}
\label{sec:il_approach}

We use behavior cloning to pretrain our
\objnav policy on the human demonstrations we collect.
Let $\pi^{BC}_{\theta}(a_t \mid o_t)$ denote a policy parametrized by $\theta$ that maps
observations $o_t$ to a distribution over actions $a_t$.
Let $\tau$ denote a trajectory consisting of state, observation, action tuples:
$\tau = \big(s_0, o_0, a_0, \ldots, s_T, o_T, a_T\big)$ and $\Tau = \big\{ \tau^{(i)} \big \}_{i=1}^N$
denote a dataset of human demonstrations.
The optimal parameters are
\begin{equation}
    \theta^* =
        \text{arg\,min}_\theta
            \sum_{i=1}^N
                \sum_{(o_{t}, a_{t}) \in \tau^{(i)}}
                    -\log \Big( \pi^{BC}_{\theta}(a_{t} \mid o_{t}) \Big)
\end{equation}
We use inflection weighting \cite{eqa_matterport} to
adjust the loss function to upweight timesteps where actions change
(\ie $a_{t-1} \neq a_t$).

Our \textbf{ObjectNav policy} architecture is a simple CNN+RNN model
from \cite{ovrl}.
To encode RGB input $(i_t =$ CNN$(I_t))$, we use a ResNet50~\cite{he_cvpr16}. Following \cite{ovrl}, the CNN is first pre-trained on the Omnidata starter dataset \cite{eftekhar2021omnidata}
using the self-supervised pretraining method DINO \cite{caron2021emerging} and then
finetuned during \objnav training.
The GPS+Compass inputs, $P_t = (\Delta x, \Delta y, \Delta z)$, and $R_t = (\Delta \theta)$,
%
are passed through fully-connected layers $p_t =$ FC$(P_t), r_t =$ FC$(R_t)$ to embed them to 32-d vectors.
Finally, we convert the object goal category to one-hot and pass it through a fully-connected layer $g_t =$ FC$(G_t)$, resulting in a 32-d vector.
All of these input features are concatenated to form an observation embedding, and fed into a 2-layer, 2048-d GRU at every timestep to predict a distribution over actions $a_t$ - formally, given current observations $o_t = [i_t, p_t, r_t, g_t]$, $(h_t, a_t) =$ GRU$(o_t, h_{t-1}) $.
To reduce overfitting, we apply color-jitter and random shifts \cite{yarats2021mastering} to the RGB inputs.


\section{RL Finetuning}
\label{sec:rl_finetuning}

%
Our motivation for RL-finetuning is two-fold. First, finetuning may allow for
higher performance as behavior cloning is known to suffer from a train/test mismatch -- when training, the policy sees the result of taking
ground-truth actions, while at test-time,
it must contend with the consequences of its own actions.
Second, collecting more human demonstrations on new scenes or simply to improve performance is time-consuming and expensive.
On the other hand,
RL-finetuning is trivially scalable (once annotated 3D scans are available) and
has the potential to reduce the amount of human demonstrations needed.

\subsection{Setup}

%
The RL objective is to find a policy $\pi_\theta(a|s)$ that
maximizes expected sum of discounted future rewards.
Let $\tau$ be a sequence of object, action, reward tuples ($o_t$, $a_t$, $r_t$) where
$a_t \sim \pi_\theta(\cdot \mid o_t)$ is the action sampled from the agent's policy,
and $r_t$ is the reward.
For a discount factor $\gamma$, the optimal policy is
\begin{equation}
    \pi^{*} = \argmax_{\pi} \mexp_{\tau \sim \pi}[R_T], \hspace{0.1em} \text{where } R_T = \sum_{t=1}^T \gamma^{t-1} r_t.
\end{equation}

To solve this maximization problem, actor-critic RL methods learn a state-value function $V(s)$ (also called a critic)
in addition to the policy (also called an actor).
The critic $V(s_t)$ represents the expected value of returns $R_t$ when starting from state $s_t$ and acting under the policy $\pi$, where returns are defined as $R_t = \sum_{i=t}^T \gamma^{i-t} r_i$.
We use DD-PPO \cite{wijmans_iclr20}, a distributed implementation
of PPO~\cite{schulman_arxiv17}, an on-policy RL algorithm.
Given a $\theta$-parameterized policy $\pi_\theta$ and a set of rollouts, PPO updates the policy as follows.
Let $\hat{A_t} = R_t - V(s_t)$, be the advantage estimate and $p_t(\theta) = \frac{\pi_\theta(a_t | o_t)}{\pi_{\theta_\text{old}}(a_t | o_t)}$ be the ratio of the probability of action $a_t$ under current policy and under the policy
used to collect rollouts.
The parameters are updated by maximizing:
\begin{equation}
    J^{PPO}(\theta) = \mexp_t \bigg[ \text{min} \big( p_t(\theta) \hat{A_t}, \text{clip}(p_t(\theta), 1 - \epsilon, 1 + \epsilon) \hat{A_t} \big) \bigg]
    \label{eq:ppo_update}
\end{equation}

We use a sparse success reward. Sparse success is simple (does not require hyperparameter optimization)
and has fewer unintended consequences (\eg Maksymets~\etal~\cite{thda_iccv21}
showed that typical dense rewards used in \objnav
actually \emph{penalize} exploration, even though
exploration is necessary for \objnav in new environments).
Sparse rewards are desirable but typically difficult to use with
RL (when initializing training from scratch) because they result
in nearly all trajectories achieving $0$ reward,
making it difficult to learn.
However, since we pretrain with BC, we do not observe any such pathologies.

\begin{figure}[h]
  \centering
    \includegraphics[width=0.98\linewidth]{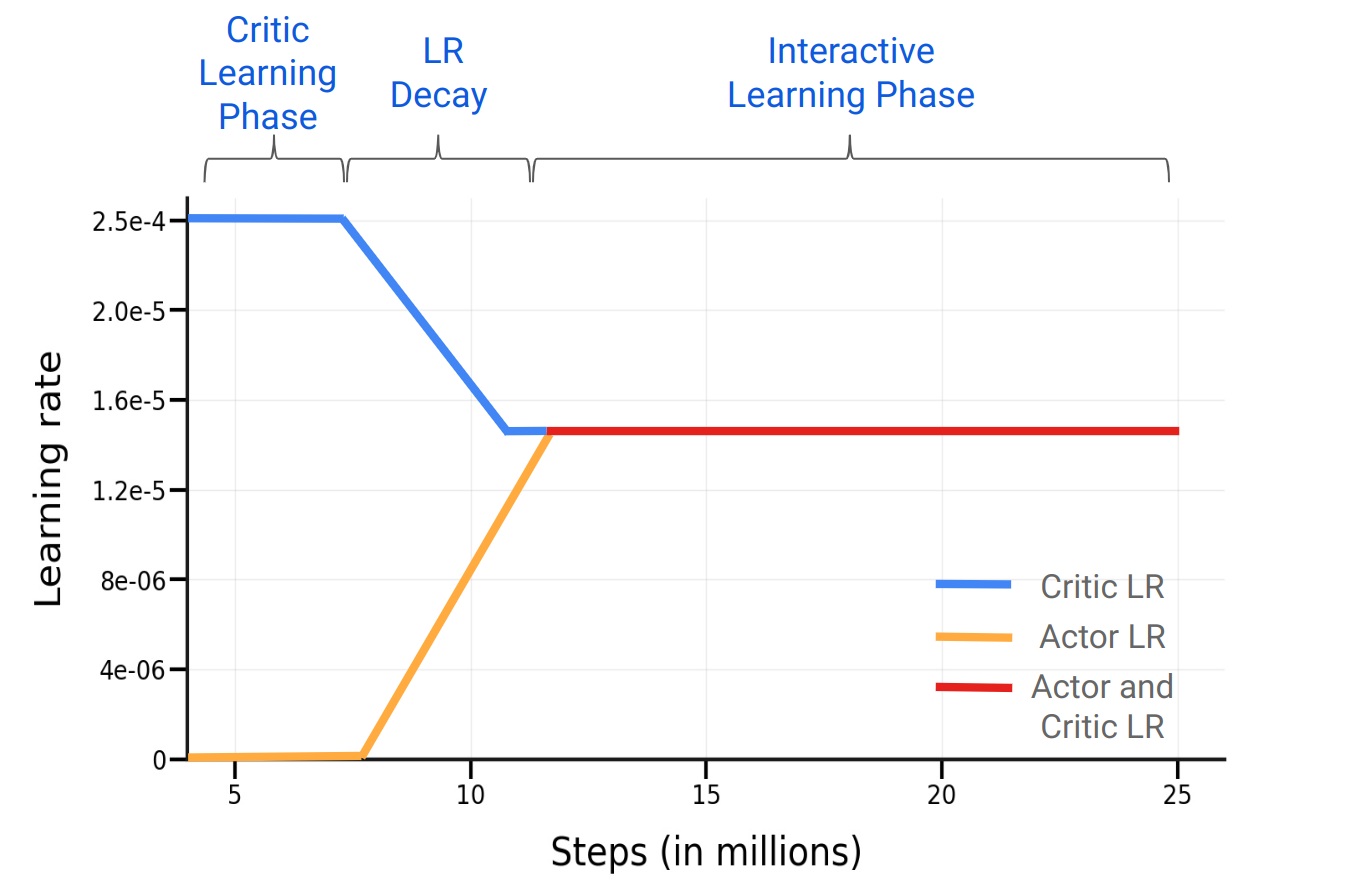}
    \caption{Learning rate schedule for RL Finetuning.}
    \vspace{-15pt}
    \label{fig:rl_ft_schedule}
\end{figure}

\subsection{Finetuning Methodology}

We use the behavior cloned policy $\pi^{BC}_{\theta}$ weights to initialize the actor parameters.
However, notice that during behavior cloning we do not
learn a critic nor is it easy to do so -- a critic learned on
human demonstrations (during behavior cloning)
would be overly optimistic since all it sees are successes.
Thus, we must learn the critic from scratch during RL.
%
Naively finetuning the actor with a randomly-initialized critic
leads to a rapid drop in performance\footnote{After the initial drop, the performance increases but the improvements on success are small.}
(see \figref{fig:rl_ft_failure}) since the critic provides poor value estimates
which influence the actor's gradient updates (see Eq.\eqref{eq:ppo_update}).
%
%
%
We address this issue by using a two-phase training regime:

\textbf{Phase 1: Critic Learning}.
%
In the first phase, we rollout trajectories using the frozen policy,
pre-trained using BC,
and use them to learn a critic.
To ensure consistency of rollouts collected for critic
learning with RL training, we sample actions (as opposed to using {\tt argmax} actions)
from the pre-trained BC policy: $a_{t} {\sim} \pi_\theta(s_{t})$.
%
%
We train the critic until its loss plateaus.
In our experiments, we found $8M$ steps to be sufficient.
In addition, we also initialize the weights of the
critic's final linear layer close to zero to stabilize training.

\textbf{Phase 2: Interactive Learning}. In the second phase, we unfreeze the actor RNN\footnote{The CNN and non-visual observation embedding layers remain frozen. We find this to be more stable.} and finetune both actor and critic weights.
We find that naively switching from phase 1 to phase 2 leads to small improvements in policy performance at convergence.
%
We gradually decay the critic learning rate
from $2.5 \times 10^{-4}$ to $1.5 \times 10^{-5}$
while warming-up the policy learning rate
from $0$ to $1.5 \times 10^{-5}$ between $8M$ to $12M$ steps,
and then keeping both at $1.5 \times 10^{-5}$ through the course of training.
See~\figref{fig:rl_ft_schedule}.
%
%
%
We find that using this learning rate schedule helps improve policy performance.
%
%
For parameters that are shared between the actor and critic (\ie the RNN),
we use the lower of the two learning rates (\ie always the actor's in our schedule).
To summarize our finetuning methodology:
\begin{compactenum}[--]
    \item First, we initialize the weights of the policy network with the IL-pretrained policy and initialize critic weights close to zero. We freeze the actor and shared weights. The only learnable parameters are in the critic.
    \item Next, we learn the critic weights on rollouts collected from the pretrained, frozen policy.
    \item After training the critic, we warmup the policy learning rate and decay the critic learning rate.
    \item Once both critic and policy learning rate reach a fixed learning rate, we train the policy to convergence.
\end{compactenum}

\vspace{-4pt}
\subsection{Results}
\vspace{-1pt}

\textbf{Comparing with the RL-finetuning approach in VPT~\cite{vpt22}}.
We start by comparing our proposed RL-finetuning approach with
the approach used in VPT~\cite{vpt22}.
Specifically, \cite{vpt22} proposed initializing the critic weights to zero,
replacing entropy term with a KL-divergence loss between the frozen IL policy and
the RL policy, and decay the KL divergence loss coefficient, $\rho$,
by a fixed factor after every iteration. Notice that this
prevents the actor from drifting too far too quickly
from the IL policy, but does not solve uninitialized critic problem.
To ensure fair comparison, we implement this method within our DD-PPO framework to ensure that any performance difference is due to the fine-tuning algorithm and not tangential implementation differences.
Complete training details are in the~\cref{sec:rl_ft_vpt}.
We keep hyperparameters constant for our approach for all experiments.
Table \ref{tab:rl_ft_comparison} reports results on \hmdval
for the two approaches using $20k$ human demonstrations.
%
We find that PIRLNav achieves $+2.2\%$ Success compared to VPT and comparable SPL.


\begin{table}[h]
    \centering
    \resizebox{0.8\columnwidth}{!}{
        \begin{tabular}{@{}lrr@{}}
            \toprule
            Method & Success $(\mathbf{\uparrow})$ & SPL $(\mathbf{\uparrow})$ \\
            \midrule
            \rownumber BC                                           & $52.0$ & $20.6$ \\
            \rownumber BC$\rightarrow$RL-FT w/ VPT     & $59.7$ {\scriptsize $\pm0.70$} & $\mathbf{28.6}$ {\scriptsize $\pm0.89$} \\
            \midrule
            \rownumber PIRLNav (Ours)                                & $\mathbf{61.9}$ {\scriptsize $\pm0.47$} & $27.9$ {\scriptsize $\pm0.56$}\\
            \bottomrule
            \end{tabular}
    }
    \vspace{5pt}
    \caption{Comparison with VPT on HM3D \textsc{val}~\cite{ramakrishnan_arxiv20,yadav2022habitat}}
    \label{tab:rl_ft_comparison}
    \vspace{-10pt}
\end{table}


\begin{table}[h]
    \centering
    \resizebox{0.95\columnwidth}{!}{
        \begin{tabular}{@{}lrr@{}}
            \toprule
            Method & Success $(\mathbf{\uparrow})$ & SPL $(\mathbf{\uparrow})$ \\
            \midrule
            \rownumber BC                                           & $52.0$ & $20.6$ \\
            \rownumber BC$\rightarrow$RL-FT                                        & $53.6$ {\scriptsize $\pm1.01$} & $\mathbf{28.6}$ {\scriptsize $\pm0.50$}  \\
            \rownumber BC$\rightarrow$RL-FT (+ Critic Learning)                            & $56.7$ {\scriptsize $\pm0.93$} & $27.7$ {\scriptsize $\pm0.82$}\\
            \rownumber BC$\rightarrow$RL-FT (+ Critic Learning, Critic Decay)              & $59.4$ {\scriptsize $\pm0.42$} & $26.9$ {\scriptsize $\pm0.38$} \\
            \rownumber BC$\rightarrow$RL-FT (+ Critic Learning, Actor Warmup)              & $58.2$ {\scriptsize $\pm0.55$} & $26.7$ {\scriptsize $\pm0.69$}\\
            \midrule
            \rownumber PIRLNav                                     & $\mathbf{61.9}$ {\scriptsize $\pm0.47$} & $27.9$ {\scriptsize $\pm0.56$} \\
            \bottomrule
            \end{tabular}
    }
    \vspace{5pt}
    \caption{RL-finetuning ablations on HM3D \textsc{val}~\cite{ramakrishnan_arxiv20,yadav2022habitat}}
    \label{tab:rl_ft_ablations}
\end{table}

\textbf{Ablations}. Next, we conduct ablation experiments to quantify the
importance of each phase in our RL-finetuning approach.
Table~\ref{tab:rl_ft_ablations} reports results on the \hmdval split for a policy
BC-pretrained on $20k$ human demonstrations and RL-finetuned for $300M$ steps,
complete training details are in~\cref{sec:rl_ft_ablations}.
First, without a gradual learning transition (row $2$),~\ie without a critic
learning and LR decay phase, the policy improves by $1.6\%$ on success and $8.0\%$ on SPL.
Next, with only a critic learning phase (row $3$), the policy improves by $4.7\%$ on success and $7.1\%$ on SPL.
Using an LR decay schedule only for the critic after the critic learning phase improves success by $7.4\%$ and SPL by $6.3\%$,
and using an LR warmup schedule for the actor (but no critic LR decay) after the critic learning phase
improves success by $6.2\%$ and SPL by $6.1\%$.
Finally, combining everything (critic-only learning, critic LR decay, actor LR warmup),
our policy improves by $9.9\%$ on success and $7.3\%$ on SPL.

\begin{table}[h] 
    \centering
    \resizebox{0.98\linewidth}{!}{
        \begin{tabular}{@{}llrrcrr@{}}
            \toprule
            & & \multicolumn{2}{c}{\textsc{test-std}} & & \multicolumn{2}{c}{\textsc{test-challenge}} \\
            \cmidrule{3-4} \cmidrule{6-7}
            & Method & Success $(\mathbf{\uparrow})$  & SPL $(\mathbf{\uparrow})$
                & & Success $(\mathbf{\uparrow})$ & SPL $(\mathbf{\uparrow})$  \\
            \midrule
            \\[-10pt]
            & \rownumber Stretch~\cite{chaplot_neurips20}
                & $60.0\%$ & $34.0\%$
                & & $56.0\%$ & $29.0\%$ \\
            & \rownumber ProcTHOR-Large~\cite{procthor}
                & $54.0\%$ & $32.0\%$
                & & - & - \\
            & \rownumber Habitat-Web~\cite{ramrakhya2022}         & $55.0\%$     & $22.0\%$ & & -           & - \\
            & \rownumber DD-PPO~\cite{habitat_challenge2022}      & $26.0\%$     & $12.0\%$ & & -           & - \\
            & \textcolor{lightgray}{\rownumber} \textcolor{lightgray}{Populus A.}
                & \textcolor{lightgray}{$66.0\%$} & \textcolor{lightgray}{$32.0\%$}
                & & \textcolor{lightgray}{$60.0\%$} & \textcolor{lightgray}{$30.0\%$} \\
            & \textcolor{lightgray}{\rownumber} \textcolor{lightgray}{ByteBOT}
                & \textcolor{lightgray}{$68.0\%$} & \textcolor{lightgray}{$37.0\%$}
                & & \textcolor{lightgray}{64.0\%} & \textcolor{lightgray}{35.0\%} \\
            \midrule
            & \rownumber PIRLNav\tablefootnote{The approach is called ``BadSeed'' on the \textsc{HM3D} leaderboard: \href{https://eval.ai/web/challenges/challenge-page/1615/leaderboard/3899}{\tt eval.ai/web/challenges/challenge-page/1615/leaderboard/3899}}
                & $\mathbf{65.0\%}$ & $33.0\%$
                & & $\mathbf{65.0}\%$ & $33.0\%$ \\
            \bottomrule
            \end{tabular}
    }
    \vspace{5pt}
    \caption{Results on \textsc{HM3D} \teststd and
    \testchallenge~\cite{habitat_challenge2022,yadav2022habitat}. Unpublished works
    submitted only to the \objnav leaderboard have been grayed out.}
    \label{tab:onav_test_std_challenge}
    \vspace{-5pt}
\end{table}

\textbf{ObjectNav Challenge 2022 Results}. Using our overall two-stage training
approach of BC-pretraining followed by RL-finetuning, we achieve state-of-the-art
results on \objnav -- $65.0\%$ success and $33.0\%$ SPL on both
the \teststd and \testchallenge splits and $70.4\%$ success and $34.1\%$ SPL on \textsc{val}.
Table~\ref{tab:onav_test_std_challenge} compares our results with the top-4
entries to the Habitat \objnav Challenge 2022 \cite{habitat_challenge2022}.
Our approach outperforms Stretch~\cite{chaplot_neurips20}
on success rate on both \teststd and \testchallenge
and is comparable on SPL ($1\%$ worse on \teststd, $4\%$ better on \testchallenge).
ProcTHOR~\cite{procthor}, which uses $10k$ procedurally-generated environments for training,
achieves $54\%$ success and $32\%$ SPL on \teststd split,
which is $11\%$ worse at success and $1\%$ worse at SPL than ours.
For sake of completeness, we also report results of two
unpublished entries uploaded to the leaderboard --
Populus A. and ByteBOT.
Unfortunately, there is no associated report yet with these entries, so we are unable to comment on the details of these approaches, or
even whether the comparison is meaningful.
%
%
\section{Role of demonstrations in BC$\rightarrow$RL transfer}
\label{sec:experiments}

Our decision to use human demonstrations for BC-pretraining before RL-finetuning
was motivated by results in prior work~\cite{ramrakhya2022}.
Next, we examine if other cheaper sources of demonstrations lead
to equally good BC$\rightarrow$RL generalization.
%
Specifically, we consider $3$ sources of demonstrations:

\textbf{Shortest paths (SP)}. These demonstrations are generated by greedily
sampling actions to fit the geodesic shortest path to the nearest navigable goal
object, computed using the ground-truth map of the environment.
%
%
These demonstrations do not capture any exploration, they only capture
success at the \objnav task via the most efficient path.\\
\textbf{Task-Agnostic Frontier Exploration (FE)~\cite{chaplot_neurips20}}.
These are generated by using a 2-stage approach: 1) Exploration: where a
task-agnostic strategy is used to maximize exploration coverage and build a top-down
semantic map of the environment, and
2) Goal navigation: once the goal object is detected by the semantic predictor,
the developed map is used to reach
it by following the shortest path. These demonstrations capture \objnav-agnostic
exploration.

\textbf{Human Demonstrations (HD)~\cite{ramrakhya2022}}.
These are collected by asking humans on Mechanical Turk to control an agent and
navigate to the goal object.
Humans are provided access to the first-person RGB view of the agent and tasked
to reach within $1$m of the goal object category.
These demonstrations capture human-like \objnav-specific exploration.

\subsection{Results with Behavior Cloning}
\label{sec:il_main}


%
Using the BC setup described in Sec.~\ref{sec:il_approach}, we train on SP,
FE, and HD demonstrations.
Since these demonstrations vary in trajectory length (\eg SP are significantly
shorter than FE), we collect ${\sim}12M$ steps of experience with each method.
That amounts to $240k$ SP, $70k$ FE, and $77k$ HD demonstrations respectively.
As shown in Table~\ref{tab:il_main},
BC on $240k$ SP demonstrations leads to $6.4\%$ success and $5.0\%$ SPL.
We believe this poor performance is due to an imitation gap~\cite{advisor},~\ie
the shortest path demonstrations are generated with access to privileged
information (ground-truth map of the environment) which is not available to the
policy during training.
Without a map, following the shortest path in a new environment to find a
goal object is not possible.
BC on $70k$ FE demonstrations achieves $44.9\%$ success and $21.5\%$ SPL,
which is significantly better than BC on shortest paths ($+38.5\%$ success, $+16.5\%$ SPL).
Finally, BC on $77k$ HD obtains the best results --
$64.1\%$ success, $27.1\%$ SPL.
%
These trends suggest that task-specific exploration (captured in human
demonstrations) leads to much better generalization than
task-agnostic exploration (FE) or shortest paths (SP).

\begin{table}[t]
    \centering
    \resizebox{0.8\columnwidth}{!}{
        \begin{tabular}{@{}lrr@{}}
            \toprule
            Training demonstrations & Success $(\mathbf{\uparrow})$ & SPL $(\mathbf{\uparrow})$ \\
            \midrule
            Shortest paths ($240k$)                                    & $6.4\%$ & $5.0\%$ \\
            Frontier exploration ($70k$)                                    & $44.9\%$ & $21.5\%$ \\
            Human demonstrations ($77k$)                                & $\mathbf{64.1}\%$ & $\mathbf{27.1}\%$ \\
            \bottomrule
            \end{tabular}
    }
    \vspace{5pt}
    \caption{Performance on HM3D \textsc{val}
        with imitation learning on SP, FE, and HD demonstrations. The size of
        each demonstration dataset is picked such that total steps of experience
        is ${\sim}12M$.}
    \label{tab:il_main}
    \vspace{-20pt}
\end{table}

\subsection{Results with RL Finetuning}

\begin{figure}[h]
\centering
    \includegraphics[width=0.6\linewidth]{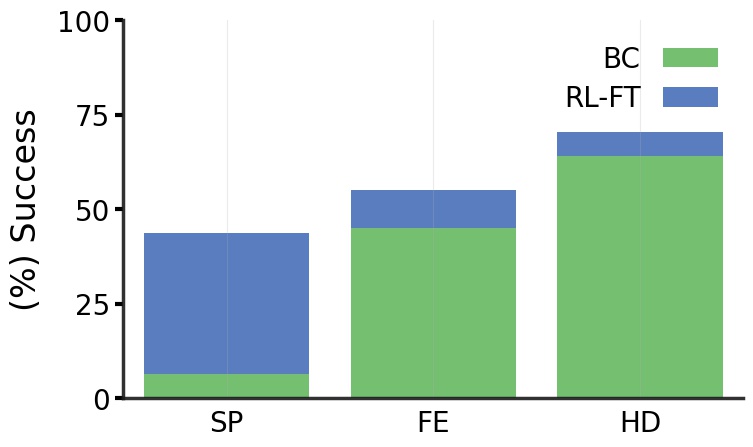}
    \caption{\objnav performance on HM3D \textsc{val}
    with BC-pretraining on shortest path (SP), frontier exploration (FE),
    and human demonstrations (HD), followed by RL-finetuning
    from each.}
    \label{fig:il_rl_best_results}
    \vspace{-15pt}
\end{figure}

Using the BC-pretrained policies on SP, FE, and HD demonstrations as initialization,
we RL-finetune each using our approach described in Sec.~\ref{sec:rl_finetuning}.
These results are summarized in \figref{fig:il_rl_best_results}.
Perhaps intuitively, the trends after RL-finetuning follow the same ordering
as BC-pretraining,~\ie RL-finetuning from BC on HD $>$ FE $>$ SP.
But there are two factors that could be leading to this ordering after
RL-finetuning --
1) inconsistency in performance at initialization (\ie BC on HD is already better
than BC on FE), and
2) amenability of each of these initializations to
RL-finetuning (\ie is RL-finetuning from HD init better than FE init?).

%

We are interested in answering (2), and so we control for (1) by selecting
BC-pretrained policy weights across SP, FE, and HD that have equal performance
on a subset of \textsc{train} $= {\sim}48.0\%$ success.
This essentially amounts to selecting BC-pretraining checkpoints for FE and HD
from earlier in training as ${\sim}48.0\%$ success is the maximum for SP.
%
%
%

\begin{figure}[t]
    \centering
    \includegraphics[width=0.98\linewidth]{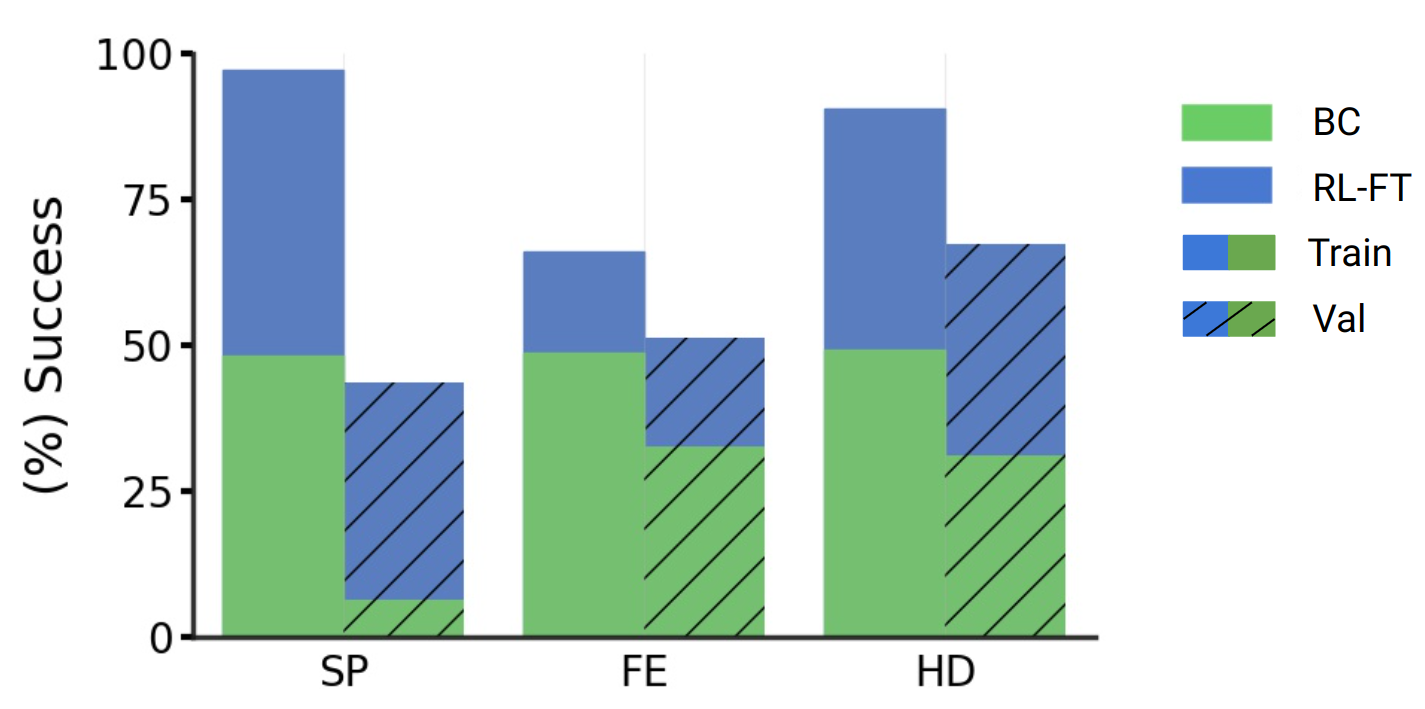}
    \caption{BC and RL performance for shortest paths (SP), frontier exploration (FE),
    and human demonstrations (HD) with equal BC training success on \textsc{HM3D} \textsc{train} (left) and \textsc{val} (right).}
    \vspace{-15pt}
    \label{fig:il_rl_main}
\end{figure}

\figref{fig:il_rl_main} shows the results after BC and RL-finetuning
on a subset of the HM3D~\textsc{train} and on \hmdval.
First, note that at BC-pretraining \textsc{train} success rates are equal ($= {\sim}48.0\%$),
while on \textsc{val} FE is slightly better than HD followed by SP.
We find that after RL-finetuning, the policy trained on HD still leads to higher
\textsc{val} success ($66.1\%$) compared to FE ($51.3\%$) and SP ($43.6\%$).
Notice that RL-finetuning from SP leads to high \textsc{train} success,
but low \textsc{val} success, indicating significant overfitting.
FE has smaller \textsc{train}-\textsc{val} gap after RL-finetuning but both
are worse than HD, indicating underfitting.
These results show that learning to imitate human demonstrations equips the
agent with navigation strategies that enable better RL-finetuning
generalization compared to imitating other kinds of demonstrations, even when
controlled for the same BC-pretraining accuracy.

\begin{table}[t]
    \centering
    \resizebox{0.9\columnwidth}{!}{
        \begin{tabular}{@{}lrr@{}}
            \toprule
            Training demonstrations &  BC Success $(\mathbf{\uparrow})$ & RL-FT Success $(\mathbf{\uparrow})$ \\
            \midrule
            \rownumber SP                                   & $\mathbf{5.2}\%$ & $34.8\%$ \\
            \rownumber HD                                   & $0.0\%$ & $\mathbf{57.2}\%$ \\
            \midrule
            \rownumber FE                                   & $\mathbf{26.3}\%$ & $43.0\%$ \\
            \rownumber HD                                   & $0.0\%$ & $\mathbf{57.2}\%$ \\
            \bottomrule
            \end{tabular}
    }
    \vspace{5pt}
    \caption{Results on SP-favoring and FE-Favoring splits.}
    \label{tab:sp_favoring_eval}
    \vspace{-15pt}
\end{table}

\textbf{Results on SP-favoring and FE-favoring episodes}.
To further emphasize that imitating human demonstrations is key to good
generalization, we created two subsplits from the \hmdval split that
are adversarial to HD performance -- SP-favoring and FE-favoring.
The SP-favoring \textsc{val} split consists of episodes where BC on SP achieved
a higher performance compared to BC on HD, \ie we select episodes
where BC on SP succeeded but BC on HD did not or
both BC on SP and BC on HD failed.
%
%
Similarly, we also create an FE-favoring \textsc{val} split using the same
sampling strategy biased towards BC on FE.
Next, we report the performance of RL-finetuned from BC on SP, FE, and HD
on these two evaluation splits in Table~\ref{tab:sp_favoring_eval}.
On both SP-favoring and FE-favoring, BC on HD is at $0\%$ success (by design), but after
RL-finetuning, is able to significantly outperform RL-finetuning from the
respective BC on SP and FE policies.
%



\subsection{Scaling laws of BC and RL}


In this section, we investigate how BC-pretraining $\rightarrow$ RL-finetuning
success scales with no. of BC demonstrations.


\textbf{Human demonstrations}.
%
We create HD subsplits ranging in size from $2k$ to $77k$ episodes,
and BC-pretrain policies with the same set of hyperparameters on each split.
Then, for each, we RL-finetune from the best-performing checkpoint.
The resulting BC and RL success on \hmdval~\vs no. of HD episodes is plotted in
\figref{fig:teaser}.
Similar to~\cite{ramrakhya2022}, we see promising scaling behavior with
more BC demonstrations.
%

Interestingly, as we increase the size of of the BC pretraining dataset and get
to high BC accuracies, the improvements from RL-finetuning decrease.
\Eg at $20k$ BC demonstrations, the BC$\rightarrow$RL improvement is $10.1\%$
success, while at $77k$ BC demonstrations, the improvement is $6.3\%$.
Furthermore, with $35k$ BC-pretraining demonstrations, the RL-finetuned
success is only $4\%$ worse than RL-finetuning from $77k$ BC demonstrations
($66.4\%$~\vs$70.4\%$).
%
%
%
Both suggest that by effectively leveraging the trade-off between
the size of the BC-pretraining dataset~\vs performance gains after RL-finetuning,
it may be possible to achieve close to state-of-the-art results without large investments
in demonstrations.

\begin{figure}
  \centering
    \includegraphics[width=0.65\linewidth]{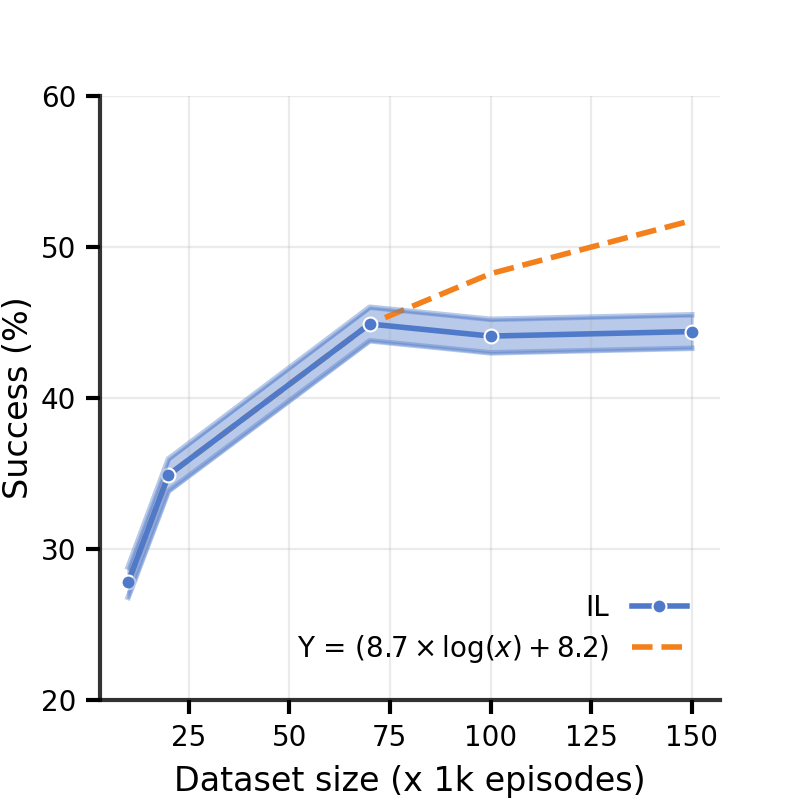}
    \vspace{5pt}
    \caption{Success on ObjectNav HM3D \textsc{val} split vs. no. of frontier
    exploration demonstrations for training.}
    \label{fig:fe_scaling}
    \vspace{-15pt}
\end{figure}

\textbf{How well does FE Scale?}
In Section~\ref{sec:il_main}, we showed that BC on human demonstrations outperforms
BC on both shortest paths and frontier exploration demonstrations, when controlled
for the same amount of training experience.
In contrast to human demonstrations however, collecting shortest paths and
frontier exploration demonstrations is cheaper, which makes scaling these
demonstration datasets easier.
Since BC performance on shortest paths is significantly worse
even with $3$x more demonstrations compared to FE and HD
($240k$ SP \vs $70k$ FE and $77k$ HD demos, Sec.~\ref{sec:il_main}),
we focus on scaling FE demonstrations.
%
%
\figref{fig:fe_scaling} plots performance on \hmdval
against FE dataset size and a curve fitted using $75k$ demonstrations to
predict performance on FE dataset-sizes $\geq$ $75k$.
We created splits ranging in size from $10k$ to $150k$.
%
Increasing the dataset size doesn't consistently improve performance and saturates
after $70k$ demonstrations, suggesting that generating more FE
demonstrations is unlikely to help.
We hypothesize that the saturation is because
these demonstrations don't capture task-specific exploration.

\vspace{-5pt}
\section{Failure Modes}
\label{sec:analysis}

\begin{figure}
  \centering
    \includegraphics[width=0.98\linewidth]{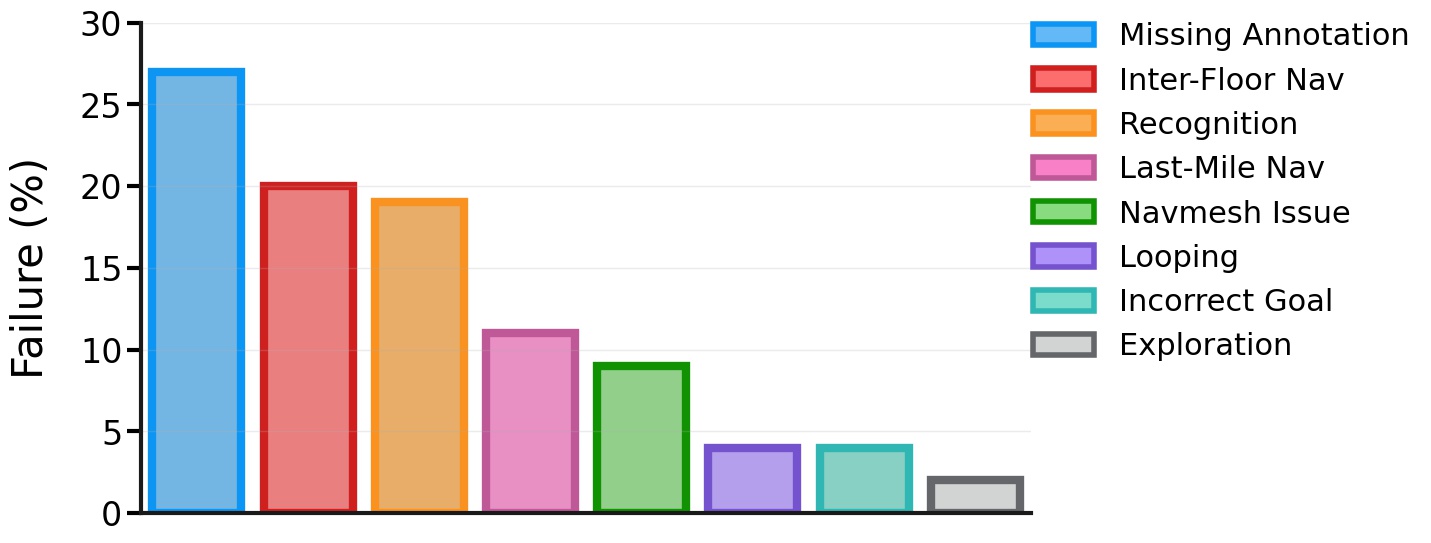}
    \caption{Failure modes of our best BC$\rightarrow$RL \objnav policy}
    \label{fig:failure_modes}
    \vspace{-15pt}
\end{figure}

To better understand the failure modes of our BC$\rightarrow$RL \objnav policies,
we manually annotate $592$ failed \hmdval episodes from our
best \objnav agent. See~\figref{fig:failure_modes}.
%
The most common failure modes are:\\
\textbf{Missing Annotations} ($27\%$): Episodes where the agent navigates
to the correct goal object category but the episode is counted
as a failure due to missing annotations in the data. \\
\textbf{Inter-Floor Navigation} ($21\%$):
The object is on a different floor and the
agent fails to climb up/down the stairs. \\
\textbf{Recognition Failure} ($20\%$): The agent sees the object in
its field of view but fails to navigate to it.\\
\textbf{Last Mile Navigation~\cite{wasserman2022lastmile}} ($12\%$).
Repeated collisions against objects or mesh geometry close to the goal object preventing the agent
from reaching close to it.\\
\textbf{Navmesh Failure} ($9\%$).
Hard-to-navigate meshes blocking the path of the agent.
\Eg in one instance, the agent fails to climb stairs because of a
narrow nav mesh on the stairs.\\
\textbf{Looping} ($4\%$).
Repeatedly visiting the same location and not exploring the rest of the environment.\\
\textbf{Semantic Confusion} ($5\%$). Confusing
the goal object with a semantically-similar object.
\Eg `armchair' for `sofa'.\\
\textbf{Exploration Failure} ($2\%$). Catch-all for failures in a
complex navigation environment, early termination, semantic failures
(\eg looking for a chair in a bathroom),~\etc.

%
As can be seen in~\figref{fig:failure_modes},
most failures (${\sim}36\%$) are due to issues in the \objnav dataset --
$27\%$ due to missing object annotations $+$
$9\%$ due to holes / issues in the navmesh.
%
%
$21\%$ failures are due to the agent being unable to climb up/down
stairs.
We believe this happens because climbing up / down stairs
to explore another floor is a difficult behavior to learn
and there are few episodes that require this.
Oversampling inter-floor navigation episodes during training
can help with this.
Another failure mode is failing to recognize the goal object --
$20\%$ where the object is in the agent's field of view but it
does not navigate to it, and $5\%$ where the agent navigates to another
semantically-similar object.
Advances in the visual backbone and object recognition can help address these.
Prior works~\cite{ramrakhya2022,chaplot_neurips20} have used explicit
semantic segmentation modules to recognize objects at each step of navigation.
Incorporating this within the BC$\rightarrow$RL training pipeline could help.
%
$11\%$ failures are due to last mile navigation, suggesting that equipping
the agent with better goal-distance estimators could help.
Finally, only ${\sim}6\%$ failures are due to looping and lack of exploration,
which is promising!

\vspace{-5pt}
\section{Conclusion}
\label{sec:conclusion}

To conclude, we propose PIRLNav, an approach to combine imitation using behavior cloning (BC) and reinforcement learning (RL)
for \objnav, wherein we pretrain a policy with BC on $77k$ human demonstrations
and then finetune it with RL, leading to state-of-the-art results on \objnav ($65\%$ success, $5\%$ improvement over previous best).
Next, using this BC$\rightarrow$RL training recipe, we present a thorough empirical study of the impact of
different demonstration datasets used for BC-pretraining on downstream RL-finetuning performance.
We show that BC / BC$\rightarrow$RL on human demonstrations outperforms BC / BC$\rightarrow$RL on shortest paths and
frontier exploration trajectories, even when we control for same BC success on \textsc{train}.
We also show that as we scale the pretraining dataset size for BC and get to higher BC success rates,
the improvements from RL-finetuning start to diminish.
%
%
Finally, we characterize our agent's failure modes, and find that
the largest sources of error are 1) dataset annotation noise, and inability of the agent
to 2) navigate across floors, and 3) recognize the correct goal object.

\xhdr{Acknowledgements}.
We thank Karmesh Yadav for OVRL model weights~\cite{ovrl},
and Theophile Gervet for answering questions related to the frontier exploration code~\cite{chaplot_neurips20}
used to generate demonstrations.
The Georgia Tech effort was supported in part by NSF, ONR YIP, and ARO PECASE. The views and conclusions contained herein are those of the authors and should not be interpreted as necessarily representing the official policies or endorsements, either expressed or implied, of the U.S. Government, or any sponsor.


\bibliographystyle{ieeetr}
\bibliography{strings,main}
\clearpage
\newpage
\newpage
\appendix

\section{Prior work in RL Finetuning}
\label{sec:prior_works}

\subsection{DAPG~\cite{rajeswaran_RSS_18}}

\textbf{Preliminaries}. Rajeswaran~\etal~\cite{rajeswaran_RSS_18} proposed DAPG, a method which incorporates demonstrations in RL, and thus quite relevant to our methodology.
DAPG first pretrains a policy using behavior cloning then finetunes the policy using an augmented RL objective (shown in Eq.~\eqref{eq:dapg}).
DAPG proposes to use different parts of demonstrations dataset during different stages of learning for tasks involving sequence of behaviors.
To do so, they add an additional term to the policy gradient objective:

\begin{multline}
    g_{aug} = \sum_{(s, a) \in \tau \sim \pi_\theta} \nabla_\theta \log_{\pi_\theta} (a|s) A^\pi(s, a) \text{ } + \\ \sum_{(s, a) \in \tau \sim { \Tau }} \nabla_\theta \log_{\pi_\theta} (a|s) w(s, a)
    \label{eq:dapg}
\end{multline}

Here $\tau \sim \pi_\theta$ is a trajectory obtained by executing the current policy, $\tau \sim \Tau$ denotes a
trajectory obtained by replaying a demonstration, and $w(s, a)$ is a weighting function to alternate between imitation and reinforcement learning.
DAPG uses a heuristic weighting scheme to set $w(s, a)$ to decay the auxiliary objective:

\begin{equation}
    w(s,a) = \lambda_0 \lambda^{k}_1 \max_{(s^{'}, a^{'}) \in \tau \sim \pi_\theta}  A^{\pi_\theta}(s^{'}, a^{'}) \forall (s, a)
    \label{eq:weight_dapg}
\end{equation}

where $\lambda_0$ and $\lambda_1$ are hyperparameters and $k$ is the update iteration counter.
The decaying weighting term $\lambda^{k}_1$ is used
to avoid biasing the gradient towards the demonstrations data towards the end of training.

\textbf{Implementation Details}. \cite{rajeswaran_RSS_18} showed results of using DAPG on dexterous hand manipulation tasks for
object relocation, in-hand manipulation, tool use,~\etc.
To train the policy with behavior cloning, they use $25$ demonstrations for each task gathered using the Mujoco HAPTIX system~\cite{mujocohaptix}.
The small size of the demonstrations dataset and the observation input allows DAPG to load the demonstrations dataset in system memory
which makes it feasible to compute the augmented RL objective shown above.

\textbf{Challenges in adopting~\cite{rajeswaran_RSS_18}'s setup}.
Compared to \cite{rajeswaran_RSS_18}, our setup uses high-dimensional visual
input (256$\times$256 RGB observations) and $77k$ \objnav demonstrations
for training.
Following DAPG's training implementation, storing the visual inputs for
$77k$ demonstrations in system memory would require $2$TB,
which is significantly higher than what is possible on today's systems.
An alternative is to leverage on-the-fly demonstration replay during RL training.
However, efficiently incorporating demonstration replay with experience collection
online requires solving a systems research problem.
Naively switching between online experience collection using the current policy
and replay demonstrations would require $2$x the current experience collection
time, overall hurting the training throughput.

\subsection{Feasibility of Off-Policy RL finetuning}

There are several methods for incorporating demonstrations with
off-policy RL~\cite{awac,awopt2021corl,kalashnikov2018scalable,AWRPeng19,marwil}.
~\cref{alg:off_policy_rl} shows the general framework of off-policy RL (finetuning) methods.
%

\begin{algorithm}
\caption{General framework of off-policy RL algorithm}
\label{alg:off_policy_rl}
\begin{algorithmic}
\Require $\pi_\theta$ : Policy, $B$: replay buffer, $N$: Rounds, $I$: Policy Update Iterations
\For{$k = 1 \text{ to } N$}
    \State Trajectory $\tau \leftarrow$ \text{Rollout } $\pi_\theta (\cdot|s)$ to collect trajectory
    $\{(s_1, a_1, r_1, h_1),......, (s_T, a_T, r_T, h_T) \}$
    \State $B \leftarrow \{B\} \cup \{\tau\}$
    \State $\pi_\theta \leftarrow $ TrainPolicy($\pi_\theta$, $B$) for $I$ iterations
\EndFor
\end{algorithmic}
\end{algorithm}

Unfortunately, most of these methods use feedforward state encoders,
which is ill-posed for partially observable settings.
In partially observable settings, the agent requires a state representation
that combines information about the state-action trajectory so far
with information about the current observation,
which is typically achieved using a recurrent network.
%

To train a recurrent policy in an off-policy setting, the full state-action trajectories need to be stored
in a replay buffer to use for training, including the hidden state $h_t$ of the RNN.
The policy update requires a sequence input for multiple time steps
$\big[(s_t, a_t, r_t, h_t),......, (s_{t+l}, a_{t+l}, r_{t+l}, h_{t+l}) \big] \sim \tau$ where $l$ is sampled sequence length.
Additionally, it is not obvious how the hidden state should be initialized for RNN updates when using a sampled sequence in the off-policy setting.
%
%
Prior work DRQN\cite{hausknecht_aaai15} compared two training strategies to train a recurrent network from replayed experience:

\begin{compactenum}
    \item \textbf{Bootstrapped Random Updates}. The episodes are sampled randomly from the replay buffer and the policy updates
    begin at random steps in an episode and proceed only for the unrolled timesteps.
    The RNN initial state is initialized to zero at the start of the update.
    Using randomly sampled experience better adheres to DQN's~\cite{dqn} random sampling strategy, but, as a result,
    the RNN's hidden state must be initialized to zero at the start of each policy update.
    Using zero start state allows for independent decorrelated sampling of short sequences which is important for robust
    optimization of neural networks.
    Although this can help RNN to learn to recover predictions from an initial state that mismatches with the hidden state
    from the collected experience but it might limit the ability of the network to rely on it's recurrent state and exploit long
    term temporal correlations.

    \item \textbf{Bootstrapped Sequential Updates}. The full episode replays are sampled randomly from the replay buffer and the
    policy updates begin at the start of the episode. The RNN hidden state is carried forward throughout the episode.
    Eventhough this approach avoids the problem of finding the correct initial state it still has computational issues
    due to varying sequence length for each episode, and algorithmic issues due to high variance of network updates
    due to highly correlated nature of the states in the trajectory.
\end{compactenum}

Even though using bootstrapped random updates with zero start states performed well in Atari which is mostly fully observable,
R2D2\cite{r2d2_iclr19} found using this strategy prevents a RNN from learning
long-term dependencies in more memory critical environments like DMLab.
\cite{r2d2_iclr19} proposed two strategies to train recurrent policies with randomly samples sequences:

\begin{compactenum}
    \item \textbf{Stored State}. In this strategy, the hidden state is stored at each step in the replay and use it to initialize
    the network at the time of policy updates.
    Using stored state partially remedies the issues with initial recurrent state mismatch in zero start state strategy but it suffers
    from `representational drfit' leading to `recurrent state staleness', as the stored state generated by a sufficiently old network
    could differ significantly from a state from the current policy.
    \item \textbf{Burn-in}. In this strategy the initial part of the replay sequence is used to unroll the network and
    produce a start state (`burn-in period') and update the network on the remaining part of the sequence.
\end{compactenum}

While R2D2~\cite{r2d2_iclr19} found a combination of these strategies to be effective at mitigating the representational drift and recurrent state staleness, this increases
computation and requires careful tuning of the replay sequence length $m$ and burn-in period $l$.

Both \cite{r2d2_iclr19,hausknecht_aaai15} demonstrate the issues associated with using a recurrent policy in an off-policy setting
and present approaches that mitigate issues to some extent.
Applying these techniques for Embodied AI tasks and off-policy RL finetuning is an open research problem and requires empirical evaluation of these
strategies.

\section{Prior work in Imitation Learning}

In Imitation Learning (IL), we use demonstrations of successful behavior to learn a policy that imitates the expert (demonstrator) providing these trajectories.
The simplest approach to IL is behavior cloning (BC), which uses supervised learning to learn a policy to imitate the demonstrator.
However, BC suffers from poor generalization to unseen states, since the training mimics the actions and not their consequences.
DAgger~\cite{ross_aistats11} mitigates this issue by iteratively aggregating the dataset using the expert and trained policy $\hat{\pi_{i-1}}$ to learn the policy $\hat{\pi_{i}}$. 
Specifically, at each step $i$, the new dataset $D_{i}$ is generated by:

\begin{equation}
    \pi_{i} = \beta \pi_{exp} + (1 - \beta) \hat{\pi}_{i-1}
\end{equation}

where, $\pi_{exp}$ is a queryable expert, and $\hat{\pi}_{i-1}$ is the trained policy at iteration $i-1$. 
Then, we aggregate the dataset $D \leftarrow D \cup D_i$ and train a new policy $\hat{\pi}_i$ on the dataset $D$.
Using experience collected by the current policy to update the policy for next iteration enables DAgger~\cite{ross_aistats11} to 
mitigate the poor generalization to unseen states caused by BC.
However, using DAgger~\cite{ross_aistats11} in our setting is not feasible as we don't have a queryable human expert for policies being trained with
human demonstrations.

Alternative approaches~\cite{ho_nips16,bahdanau_iclr19,abbeel_icml04,max_ent_irl,fu2018learning} for imitation learning are variants of inverse reinforcement learning (IRL), which learn reward function from expert demonstrations in order to train a policy. 
IRL methods learn a parameterized $\mathcal{R_{\phi}(\tau )}$ reward function, which models the behavior of the expert and assigns a scalar reward to a demonstration.
Given the reward $r_t$, a policy $\pi_\theta(a_t|s_t)$ is learned to map states $s_t$ to distribution over actions $a_t$ at each time step.
The goal of IRL methods is to learn a reward function such that a policy trained to maximize the discounted sum of the learned reward matches the behavior of the demonstrator.
Compared to prior works~\cite{ho_nips16,bahdanau_iclr19,abbeel_icml04,max_ent_irl,fu2018learning}, our setup uses a partially-observable setting and high-dimensional visual input for training.
Following training implementation from prior works, storing visual inputs of demonstrations for reward model training would require $2TB$ system memory, which is significantly higher than what is possible on today's systems.
Alternatively, efficiently replaying demonstrations during RL training with reward model learning in the loop requires solving an open systems research problem.
%
In addition, applying these methods for tasks in a partially observable setting is an open research problem and requires empirical evaluation of these approaches.

\section{Training Details}
\vspace{-5pt}

\subsection{Behavior Cloning}

We use a distributed implementation of behavior cloning by~\cite{ramrakhya2022} for our imitation pretraining.
Each worker collects $64$ frames of experience from $8$ environments parallely by replaying actions from the demonstrations
dataset.
We then perform a policy update using supervised learning on $2$ mini batches.
For all of our BC experiments, we train the policy for $500M$ steps on $64$ GPUs using Adam optimizer
with a learning rate $1.0\times10^{-3}$ which is linearly decayed after each policy update.
\cref{tab:il_hyperparameters} details the default hyperparameters used in all of our training runs.

\begin{table}[t]
    \centering
        \begin{tabular}{@{}lr@{}}
            \toprule
            Parameter &  Value \\
            \midrule
            Number of GPUs & 64 \\
            Number of environments per GPU & 8 \\
            Rollout length & 64 \\
            Number of mini-batches per epoch & 2 \\
            Optimizer & Adam \\
            \quad{}Learning rate & $1.0\times10^{-3}$ \\
            \quad{}Weight decay & $0.0$ \\
            \quad{}Epsilon & $1.0\times10^{-5}$ \\
            DDPIL sync fraction & 0.6 \\
            \bottomrule
            \end{tabular}
    \vspace{5pt}
    \caption{Hyperparameters used for Imitation Learning.}
    \label{tab:il_hyperparameters}
\end{table}

\subsection{Reinforcement Learning}

To train our policy using RL we use PPO with Generalized Advantage Estimation (GAE) \cite{schulman_iclr16}.
We use a discount factor $\gamma$ of $0.99$ and set GAE parameter $\tau$ to 0.95.
We do not use normalized advantages.
To parallelize training, we use DD-PPO with $16$ workers on $16$ GPUs.
Each worker collects $64$ frames of experience from $8$ environments parallely and then performs
$2$ epochs of PPO update with $2$ mini batches in each epoch.
For all of our experiments, we RL finetune the policy for $300M$ steps.
\cref{tab:rl_hyperparameters} details the default hyperparameters used in all of our training runs.

\begin{table}[t]
    \centering
        \begin{tabular}{@{}lr@{}}
            \toprule
            Parameter &  Value \\
            \midrule
            Number of GPUs & 16 \\
            Number of environments per GPU & 8 \\
            Rollout length & 64 \\
            PPO epochs & 2 \\
            Number of mini-batches per epoch & 2 \\
            Optimizer & Adam \\
            \quad{}Weight decay & $0.0$ \\
            \quad{}Epsilon & $1.0\times10^{-5}$ \\
            PPO clip & 0.2 \\
            Generalized advantage estimation & True \\
            \quad{}$\gamma$ & 0.99 \\
            \quad{}$\tau$ & 0.95 \\
            Value loss coefficient & 0.5 \\
            Max gradient norm & 0.2 \\
            DDPPO sync fraction & 0.6 \\
            \bottomrule
            \end{tabular}
    \vspace{5pt}
    \caption{Hyperparameters used for RL finetuning.}
    \label{tab:rl_hyperparameters}
\end{table}

\subsection{RL Finetuning using VPT}
\label{sec:rl_ft_vpt}

To compare with RL finetuning approach proposed in VPT \cite{vpt22} we implement the method in DD-PPO framework.
Specifically, we initialize the critic weights to zero, replace the entropy term in PPO~\cite{schulman_arxiv17} with a
KL-divergence loss between the frozen IL policy and RL policy, and decay the KL divergence loss coefficient, $\rho$,
by a fixed factor after every iteration. This loss term is defined as:

\begin{equation}
    L_{kl\_penalty} =  \rho \text{KL} (\pi^{BC}_{\theta}, \pi_\theta)
\end{equation}

where $\pi^{BC}_{\theta}$ is the frozen behavior cloned policy, $\pi_\theta$ is the current policy, and $\rho$ is the loss weighting term.
Following, VPT~\cite{vpt22} we set $\rho$ to $0.2$ at the start of training and decay it by $0.995$ after each policy update.
We use learning rate of $1.5\times10^{-5}$ without a learning rate decay for our VPT~\cite{vpt22} finetuning experiments.

\subsection{RL Finetuning Ablations}
\label{sec:rl_ft_ablations}

\begin{figure}[h]
  \centering
    \includegraphics[width=0.98\linewidth]{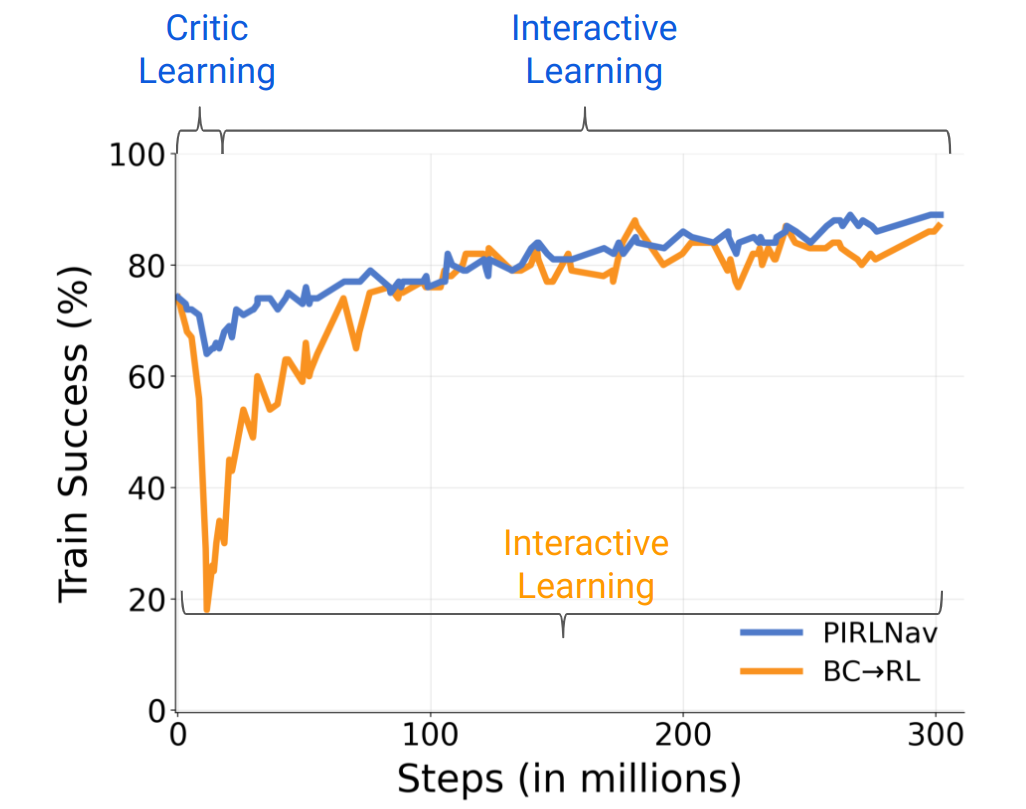}
    \caption{A policy pretrained on the \objnav task is used as initialization for
    actor weights and critic weights are initialized randomly
    for RL finetuning using DD-PPO. The policy performance
    immediately starts dropping early on during training
    and then recovers leading to slightly higher performance with further training.}
    \label{fig:rl_ft_failure}
\end{figure}


\begin{table}[h]
    \centering
    \resizebox{0.95\columnwidth}{!}{
        \begin{tabular}{@{}lrr@{}}
            \toprule
            Method & Success $(\mathbf{\uparrow})$ & SPL $(\mathbf{\uparrow})$ \\
            \midrule
            \rownumber BC                                           & $52.0$ & $20.6$ \\
            \rownumber BC$\rightarrow$RL-FT                                        & $53.6$ {\scriptsize $\pm1.01$} & $\mathbf{28.6}$ {\scriptsize $\pm0.50$}  \\
            \rownumber BC$\rightarrow$RL-FT (+ Critic Learning)                            & $56.7$ {\scriptsize $\pm0.93$} & $27.7$ {\scriptsize $\pm0.82$}\\
            \rownumber BC$\rightarrow$RL-FT (+ Critic Learning, Critic Decay)              & $59.4$ {\scriptsize $\pm0.42$} & $26.9$ {\scriptsize $\pm0.38$} \\
            \rownumber BC$\rightarrow$RL-FT (+ Critic Learning, Actor Warmup)              & $58.2$ {\scriptsize $\pm0.55$} & $26.7$ {\scriptsize $\pm0.69$}\\
            \midrule
            \rownumber PIRLNav                                     & $\mathbf{61.9}$ {\scriptsize $\pm0.47$} & $27.9$ {\scriptsize $\pm0.56$} \\
            \bottomrule
            \end{tabular}
    }
    \vspace{5pt}
    \caption{RL-finetuning ablations on HM3D \textsc{val}~\cite{ramakrishnan_arxiv20,yadav2022habitat}}
    \label{tab:rl_ft_ablations_app}
\end{table}

For ablations presented in Sec. 4.3 of the main paper (also
shown in ~\cref{tab:rl_ft_ablations_app})
we use a policy pretrained on $20k$ human demonstrations using BC and
finetuned for $300M$ steps using hyperparameters from~\cref{tab:rl_hyperparameters}.
We try $3$ learning rates ($1.5 \times 10^{-4}$, $2.5 \times 10^{-4}$, and $1.5 \times 10^{-5}$)
for both BC $\rightarrow$ RL (row 2) and BC $\rightarrow$ RL (+ Critic Learning) (row 3)
and we report the results with the one that works the best.
For PIRLNav we use a starting learning rate of $2.5 \times 10^{-4}$ and decay it to $1.5 \times 10^{-5}$, consistent with
learning rate schedule of our best performing agent.
For ablations we do not tune learning rate parameters of PIRLNav, we hypothesize tuning the parameters would help
improve performance.

We find BC $\rightarrow$ RL (row 2) works
best with a smaller learning rate but the training performance drops significantly early
on, due to the critic providing poor value estimates,
and recovers later as the critic improves. See \figref{fig:rl_ft_failure}.
In contrast when using proposed two phase learning setup with the learning rate schedule we do not observe
a significant drop in training performance.



\end{document}